\documentclass{article}

\usepackage[preprint]{icml2026}
\usepackage[utf8]{inputenc}
\usepackage[T1]{fontenc}
\usepackage{hyperref}
\usepackage{url}
\usepackage{booktabs}
\usepackage{amsfonts}
\usepackage{nicefrac}
\usepackage{microtype}
\usepackage{graphicx}
\usepackage{amsmath}
\usepackage{amssymb}
\usepackage{mathtools}
\usepackage{algorithm}
\usepackage{algorithmic}
\usepackage{amsthm}
\usepackage{tikz}
\usetikzlibrary{shapes.geometric, arrows.meta, positioning, fit, backgrounds, calc, shadows.blur}

\usepackage[most]{tcolorbox}
\usepackage{listings}

\newtcolorbox{casestudybox}{
    colback=gray!5,
    colframe=gray!50!black,
    fonttitle=\bfseries,
    title=Appendix: PACER Case Study,
    enhanced,
    breakable,
    attach title to upper,
    after title={\par\medskip},
    coltitle=black,
    fontupper=\small\ttfamily,
}
\theoremstyle{plain}      % bold title, italic body
\newtheorem{proposition}{Proposition}

\newtheorem{corollary}{Corollary}

\theoremstyle{definition} % bold title, upright body
\newtheorem{assumption}{Assumption}

\theoremstyle{remark}     % italic title, upright body

\newcommand{\packet}{\mathcal{P}}

\icmltitlerunning{A Single Revision Step Improves LLM Reasoning}

\begin{document}

\twocolumn[
\icmltitle{A Single Revision Step Improves Token-Efficient LLM Reasoning}

% 匿名化作者信息
\icmlsetsymbol{equal}{*}
\begin{icmlauthorlist}
\icmlauthor{Yingchuan Zhang}{uga}
\icmlauthor{Terry Ma}{cmu}
\icmlauthor{Wenxuan Zhong}{uga}
\icmlauthor{Ping Ma}{uga}

\icmlaffiliation{cmu}{School of Computer Science, Carnegie Mellon University, Pittsburgh, PA, USA}
\icmlaffiliation{uga}{Department of Statistics, University of Georgia, Athens, GA, USA}
\end{icmlauthorlist}

\icmlkeywords{Machine Learning, Large Language Models, Inference Efficiency, Consensus}

\vskip 0.3in
]

\begin{abstract} Large language models (LLMs) achieve higher accuracy on challenging reasoning tasks by scaling test-time compute through multiple trajectory sampling. However, standard aggregation methods like majority voting or individual confidence-based filtering face a fundamental "blind spot": they evaluate each trace in isolation. As problems scale in difficulty, models often generate hallucinated paths that exhibit misleadingly high confidence, causing the true solution to be suppressed by a narrow margin in traditional voting. We ask: can we enable traces to "peer-review" each other to resolve these near-miss errors?

We introduce 
\textbf{Packet-Conditioned Revision (PACER)}, a training-free, inference-only framework that enables reasoning traces to revise their conclusions through a structured coordination step. After a preliminary screening of generated traces, PACER constructs a compact \emph{consensus packet} containing (i) unique candidate answers, (ii) their aggregated confidence scores, and (iii) representative reasoning summaries for each candidate answer. Individual traces then perform a targeted self-review conditioned on this packet, allowing them to identify specific logical junctions where they diverged from the broader consensus and pivot if their original reasoning is found to be flawed. Final predictions are obtained via confidence-weighted voting over these revised trajectories. On challenging competitive math benchmarks such as AIME and BRUMO, PACER matches or exceeds the accuracy of 256-sample majority voting, significantly outperforming raw ensemble baselines by transforming simple consensus into a collaborative logical refinement process. \end{abstract}
% \section{Introduction}
\section{Introduction}
\begin{figure*}[t]
\centering
\includegraphics[width=0.9\textwidth]{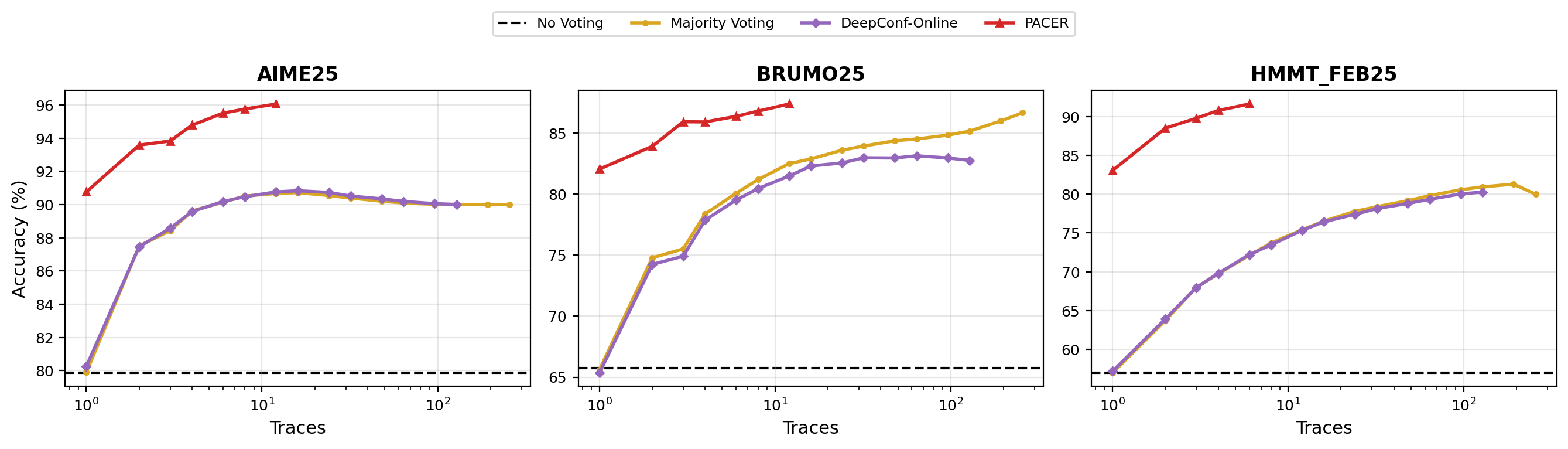}
\caption{\textbf{Token-Accuracy Pareto Frontiers with GPT OSS 120B.} We evaluate performance scaling across varying sample budgets on AIME 2025, BRUMO 2025, and HMMT Feb 2025. \textbf{PACER (red)} consistently dominates the Pareto frontier, significantly outperforming both standard Majority Voting (yellow) and DeepConf-Online (purple)}
\label{fig:accuracy_vs_budget}
\label{fig:accuracy_vs_budget}
\vspace{-10pt}
\end{figure*}

Large language models (LLMs) can solve substantially harder problems when we allocate more computation at inference time.
A particularly effective family of techniques samples multiple reasoning traces (a.k.a.\ chains-of-thought or solution trajectories) and aggregates their final answers.
Majority voting (MV), also called self-consistency, instantiates this idea by generating many independent traces and returning the most frequent answer \cite{wang2022self}.
This simple ensemble often improves accuracy on challenging reasoning benchmarks, including math and logic tasks where single-sample decoding is brittle \cite{wei2022chain}.

\textbf{The problem: accuracy gains are tied to expensive, redundant tokens.}
The main drawback of MV is cost.
Generating $N$ full traces costs roughly $N$ times the tokens (and latency) of a single run, which quickly becomes prohibitive in real deployments.
Moreover, MV treats all traces as equally informative: a trace that is internally inconsistent, hallucinated, or off-distribution contributes one full vote just like a coherent trace.
This is especially wasteful because many errors manifest early in decoding (e.g., unstable token probabilities, oscillating intermediate values), yet MV pays for the entire completion before it can discard the result.

\textbf{Why existing efficiency methods are still limited.}
Recent work improves \emph{token efficiency} by allocating compute adaptively rather than uniformly.
DeepConf \cite{fu2025deep} monitors token-level uncertainty during decoding and early-stops trajectories that appear unreliable, substantially reducing wasted generation.
Adaptive stopping rules based on answer-frequency estimates provide complementary routes to efficiency \cite{hou2024adaptive}.
However, these methods largely preserve a key structural assumption of self-consistency: \emph{traces are generated and evaluated in isolation.}
Once a trace commits to an answer, that trace does not get to reconsider after observing what other traces concluded.
% This independence is not just an implementation detail—it removes a natural error-correction channel.
This independence is especially problematic in hard math problems where a model can be 'confidently wrong,' generating high-confidence hallucinations that pass through existing filters like DeepConf but could be easily identified when contrasted with alternative logical steps.
In practice, ensembles frequently contain (i) a plurality of mutually consistent but wrong answers supported by similar flawed reasoning patterns, and (ii) a smaller set of correct answers supported by a qualitatively different rationale.
If traces cannot condition on set-level evidence, the system must rely on the raw confidence signal alone, which can be fragile under tight token budgets.

\textbf{A coordination gap at test time.}
There is thus a gap between two extremes:
(1) \emph{Self-consistency/MV} aggregates many full traces and benefits from diversity, but pays a large token cost \cite{wang2022self};
(2) \emph{Confidence-guided early stopping} reduces cost, but produces a smaller, screened set of traces that are still treated as independent, leaving potential accuracy improvements unrealized \cite{fu2025deep}.
A natural response is to let traces interact through refinement or debate.
But iterative refinement methods typically re-check a \emph{single} trace without access to peer evidence (e.g., self-refinement) or require multiple long rounds of interaction (e.g., debate), increasing latency and tokens \cite{madaan2024self, shinn2024reflexion, du2023improving, liang2023encouraging}.
This motivates a focused question:
\emph{Can we introduce a minimal, training-free coordination step that enables revision using peer evidence, while preserving the token savings of confidence-guided sampling?}

\textbf{Our approach: a single revision step with low-bandwidth peer evidence.}
We propose \textbf{Packet-Conditioned Revision (PACER)}, a lightweight coordination layer that sits on top of token-efficient sampling (we instantiate it with DeepConf-Online \cite{fu2025deep}).
PACER operates in three stages:
(1) generate a pool of \emph{completed, screened} traces under an inference budget;
(2) summarize the pool into a compact \emph{consensus packet} that contains only low-bandwidth set-level evidence—top distinct answers, their support counts, and a short representative rationale per answer;
(3) prompt each trace to perform a brief self-review conditioned on this packet and optionally revise its final answer.
The final prediction is then obtained by confidence-weighted vote over the revised answers.
Intuitively, the packet provides just enough information for a trace to notice when its conclusion is weakly supported relative to alternatives, or when another answer is backed by a more coherent rationale—without exposing the model to the full (token-heavy) set of traces.

\textbf{How PACER differs from prior work.}
PACER is \emph{not} another way to stop early; it is a post-hoc coordination step that can be appended to existing pipelines.
% Compared to DeepConf \cite{fu2025deep}, PACER adds a single round of revision that explicitly conditions on set-level evidence, addressing the isolation limitation.
Compared to DeepConf \cite{fu2025deep}, PACER does not compete as a stopping rule; instead, it targets the 'residual errors'—traces that successfully 'fool' the stopping criterion with high local confidence but collapse under global peer scrutiny. By introducing peer evidence, PACER enables these high-confidence but incorrect traces to self-correct.
Compared to self-consistency/MV \cite{wang2022self}, PACER aims to recover ensemble-like robustness under a smaller effective pool by enabling \emph{repair} of wrong traces rather than merely counting them.
Compared to self-refinement methods \cite{madaan2024self, shinn2024reflexion}, PACER's  novelty is that revision is driven by a structured summary of \emph{peer} outcomes (counts + representative rationales), not only by introspection on one’s own trace.
Compared to multi-agent debate \cite{du2023improving, liang2023encouraging}, PACER uses a single short revision step (no multi-round interaction), keeping overhead small and predictable.

\textbf{Contributions.}
Our contributions are:
\begin{enumerate}
\item \textbf{Coordination primitive:} We formalize the \emph{consensus packet}, a compact representation of an ensemble that exposes answer support counts and representative rationales, enabling set-level conditioning with overhead that depends on $N$ (top answers) rather than the number of traces.
\item \textbf{Training-free single-round revision:} We introduce PACER, an inference-only procedure that plugs into confidence-guided sampling (e.g., DeepConf-Online) and performs a single short consensus-conditioned self-review per trace.
\item \textbf{Theory:} We provide a simple repair-vs-damage condition under which a revision step improves expected trace accuracy (Section~\ref{sec:theory}), clarifying when consensus conditioning is beneficial.
\item \textbf{Measurable gains under a fixed attempt budget:} On competitive math benchmarks (AIME 2024/2025, HMMT 2025, BRUMO 2025), PACER consistently improves the accuracy--token tradeoff, matching or exceeding MV@256 while using far fewer generated tokens. For example, on \textbf{HMMT 2025}, PACER with GPT-OSS improves over the strong DeepConf-Online baseline by \textbf{+10.0} absolute percentage points (28/30 vs 25/30) (Table~\ref{tab:main_results}).
\end{enumerate}

\section{Related Work}

We review prior work on (i) \emph{parallel} test-time scaling via sampling-and-aggregation, (ii) \emph{token-efficient} variants that reduce wasted inference compute, and (iii) \emph{sequential} refinement/interaction methods. We then position PACER as a lightweight coordination step that bridges parallel scaling with a single round of revision, without training.

\textbf{Parallel test-time scaling via sampling and aggregation.}
A standard way to improve LLM reasoning at inference time is to sample multiple solution traces and aggregate them.
Self-consistency / majority voting (MV) \cite{wang2022self} is a widely used instantiation: it reduces stochastic errors by returning the most frequent extracted answer across sampled chains of thought.
Related approaches include best-of-$N$ selection with verifiers or reward models \cite{cobbe2021training}, as well as search-based decoding procedures (e.g., tree search over intermediate thoughts) that expand multiple branches under a fixed model \cite{yao2023tree}.
These methods can deliver large accuracy gains, but their cost typically scales with the number (and length) of sampled traces.

\textbf{Token-efficient parallel scaling and adaptive stopping.}
A growing line of work targets the inefficiency of fixed-size ensembles by \emph{adapting} either (a) how many samples to draw, or (b) how long to let each trace unfold.
Adaptive-consistency-style methods stop sampling once the leading answer appears sufficiently dominant \cite{hou2024adaptive}.
Early-stopping self-consistency (ESC) serializes sampling into windows and halts once the answer distribution concentrates, reducing the number of required samples \cite{li2024esc}.
More recently, methods exploit \emph{confidence/uncertainty signals} to decide whether to continue: DeepConf monitors token-level uncertainty and terminates low-quality trajectories early \cite{fu2025deep};
CGES frames stopping as a Bayesian decision rule using scalar confidence signals \cite{aghazadeh2025cges};
ES-CoT stops within a single chain by detecting convergence of intermediate ``step answers'' \cite{mao2025escot};
and entropy-based signals have also been proposed for early stopping in reasoning traces \cite{sharma2025thinkjustenough}.
Complementary empirical studies analyze test-time scaling and compute-optimal inference trade-offs across model size and inference strategy \cite{wu2024inference_scaling, liu2025compute_optimal_tts}.
\emph{Limitation shared by these approaches:} even when they adaptively allocate compute, they still generate and evaluate traces largely \emph{in isolation}, without allowing a trace to revise after seeing what the \emph{set} of traces supports.

\textbf{Set-level aggregation beyond voting.}
Several works go beyond heuristic voting by making aggregation itself more expressive.
Universal Self-Consistency (USC) prompts an LLM to \emph{select} the most consistent candidate among multiple samples, avoiding brittle answer-extraction heuristics \cite{chen2023usc}.
Sample Set Aggregator (SSA) trains a separate compact model to reason over multiple parallel samples and output a final answer \cite{qi2025ssa}, improving aggregation at the cost of additional training.
Other directions add external evaluators/verifiers and scale their number or diversity (e.g., multi-agent verification) \cite{lifshitz2025mav}.
\emph{PACER differs} in that it remains \textbf{training-free} and uses no additional verifier model; instead, it injects a minimal coordination signal (a compact consensus packet) and performs only a \emph{single} revision step per trace before a final vote.

\textbf{Sequential refinement, self-critique, and multi-agent interaction.}
Sequential test-time scaling improves an initial output through iterative revision and feedback.
Self-Refine alternates between generating feedback and refining the answer, repeatedly \cite{madaan2023selfrefine};
Reflexion uses verbal feedback stored in memory to improve subsequent trials in agentic settings \cite{shinn2023reflexion};
and multi-agent debate frameworks exchange arguments over multiple rounds to improve factuality and reasoning \cite{du2023improving, liang2023encouraging}.
These methods can be powerful but often incur multiple long rounds of generation, increasing latency and token cost.

\textbf{Positioning PACER.}
PACER occupies a middle ground between parallel ensembling and sequential refinement.
Like self-consistency and DeepConf-style screening, PACER starts from a pool of sampled traces; unlike them, PACER adds \textbf{one} low-token coordination round by summarizing the pool into a consensus packet (answers, counts, representative rationales) and letting each trace \emph{revise} conditioned on that shared evidence.
Compared to self-refinement, PACER's  revision is \emph{consensus-conditioned} (set-aware) rather than purely self-contained; compared to USC/SSA, PACER does not replace aggregation with a trained selector/aggregator, but instead improves the trace set \emph{before} a simple majority vote.

\section{Preliminaries and Problem Formulation}
\label{sec:prelim}

We study training-free test-time scaling for reasoning tasks under a fixed inference-token budget.
Let $x$ be a prompt (e.g., a math problem) and let an LLM with parameters $\theta$ generate a token sequence (a reasoning trace) $\tau=(w_1,\ldots,w_T)$.
We assume an \emph{answer extractor} $\textsc{Extract}(\tau)\mapsto a$ that maps a completed trace to a discrete final answer $a$ (e.g., by parsing the content inside \texttt{\textbackslash boxed\{\}}).
We write $a=\textsc{Extract}(\tau)$ and denote the ground-truth answer by $y^*$.

\textbf{Inference-time sampling and aggregation.}
A standard approach samples multiple traces $\{\tau_i\}_{i=1}^n$ independently and aggregates extracted answers $\{a_i\}_{i=1}^n$ by majority vote:
\vskip -0.14in
\begin{equation}
\hat{a}_{\text{MV}}
=\arg\max_a \sum_{i=1}^n \mathbb{I}[a_i=a].
\label{eq:mv}
\end{equation}
\vskip -0.14in
While effective, MV requires generating many full traces, so its token cost scales approximately linearly with $n$.

\subsection{Token-Level Uncertainty Signals}
\label{subsec:uncertainty}

During decoding, the model produces a next-token distribution $P_\theta(v\mid x,w_{<t})$ at step $t$.
Following confidence-guided sampling methods \cite{fu2025deep}, we use token-level uncertainty as a proxy for trace instability.
Concretely, we define a top-$k$ negative log-probability statistic
\vskip -0.184in
\begin{equation}
U_t \;=\; -\frac{1}{k}\sum_{j=1}^{k}\log P_\theta\!\left(v_t^{(j)} \mid x,w_{<t}\right),
\label{eq:topk_uncert}
\end{equation}
\vskip -0.184in
where $v_t^{(j)}$ is the $j$-th most likely token at step $t$.
Larger $U_t$ indicates greater local uncertainty (less stable decoding).

To reduce noise, we use a windowed average with window size $W$:
\begin{equation}
\bar{U}_t \;=\; \frac{1}{W}\sum_{r=\max(1,t-W+1)}^{t} U_r.
\label{eq:window_uncert}
\end{equation}
\subsection{A Reproducible DeepConf-Online Variant}
\label{subsec:deepconf_variant}

DeepConf-style online early stopping terminates a trace once its uncertainty becomes unusually high for the current prompt \cite{fu2025deep}.
For reproducibility, we specify a concrete stopping rule based on the \emph{prefix maximum} of windowed uncertainty.

\textbf{Prefix instability and stability.}
Define the \emph{prefix instability} up to step $t$ as
\begin{equation}
I_t(\tau) \;=\; \max_{1\le r\le t}\bar{U}_r,
\label{eq:prefix_instab}
\end{equation}
and define the corresponding \emph{prefix stability} as
\begin{equation}
S_t(\tau) \;=\; -I_t(\tau).
\label{eq:prefix_stab}
\end{equation}
By construction, $S_t(\tau)$ is non-increasing in $t$ (once a high-uncertainty region appears, the maximum cannot decrease).
For a completed trace $\tau$, we write $S(\tau)=S_T(\tau)$.

\textbf{Warmup threshold estimation.}
We first sample $N_{\text{init}}$ \emph{full} warmup traces without early stopping:
$\mathcal{T}_{\text{warm}}=\{\tau_j\}_{j=1}^{N_{\text{init}}}$.
We compute $S(\tau_j)$ for each warmup trace and set a screening threshold as a tail percentile:
\begin{equation}
s \;=\; \textsc{Percentile}\Big(\{S(\tau_j)\}_{j=1}^{N_{\text{init}}},\ 100-\eta\Big),
\label{eq:threshold}
\end{equation}
so that ``Top-$\eta\%$'' screening corresponds to keeping traces with $S(\tau)\ge s$.

\textbf{Online early stopping.}
For each subsequent sampling attempt, we decode token-by-token and maintain $S_t(\tau)$ online.
We \emph{early-stop and discard} the attempt at the first time $t$ such that
\begin{equation}
S_t(\tau) < s.
\label{eq:early_stop_rule}
\end{equation}
Because $S_t$ is non-increasing, once the trace falls below $s$ it cannot recover, making this rule compute-efficient.

\textbf{Budgets and terminology.}
To avoid ambiguity between ``attempts'' and ``completed traces,'' we distinguish:
\begin{itemize}
\item $N_{\text{try}}$: the number of sampling \emph{attempts} (some attempts may be early-stopped).
\item $B$: the number of \emph{completed traces} retained for downstream aggregation/coordination.
\end{itemize}
In our main setting, we fix $N_{\text{try}}$ (e.g., $256$ total attempts including warmup) and let the number of completed online traces be data-dependent.

\subsection{Stable Pool Construction and Answer Extraction}
\label{subsec:pool_extraction}

Let $\mathcal{T}_{\text{surv}}$ be the set of \emph{completed} online traces that were not early-stopped.
PACER coordinates over a stable pool of completed traces:
\begin{equation}
\mathcal{T}
\;=\;
\{\tau\in\mathcal{T}_{\text{warm}} \mid S(\tau)\ge s\}
\;\cup\;
\mathcal{T}_{\text{surv}}.
\label{eq:stable_pool_prelim}
\end{equation}
For each $\tau\in\mathcal{T}$ we extract an answer $a=\textsc{Extract}(\tau)$ and form pairs $(\tau,a)$.

\textbf{Practical answer extraction.}
In all experiments we prompt the model to place the final answer in \texttt{\textbackslash boxed\{\}} and define $\textsc{Extract}(\tau)$ as the substring inside the last occurrence of \texttt{\textbackslash boxed\{\}}.
If parsing fails, we fall back to a simple heuristic (e.g., last numeric token span).
We report extraction-failure rates in the appendix when applicable.

\textbf{Top-$N$ Candidate Selection.} Given the stable pool $\mathcal{T}$, we aggregate extracted answers using confidence-weighted sums: $V(a) = \sum_{(\tau,a') \in \mathcal{T}, a'=a} S(\tau)$. We then select the top $N$ answers $\mathcal{A}_{\text{top}} = \{a^{(1)}, \dots, a^{(N)}\}$ with the highest $V(a)$ as candidates for the subsequent coordination phase.

\textbf{Representative Trace Selection.} For each candidate answer $a \in \mathcal{A}_{\text{top}}$, we identify a single \emph{representative trace} $\tau^*_a$ from the stable pool that provides the most stable justification for that answer. We select the trace with the maximum prefix stability:
\begin{equation}
\tau^*_a \;=\; \arg\max_{\{\tau \in \mathcal{T} \mid \textsc{Extract}(\tau)=a\}} S(\tau).
\label{eq:rep_trace}
\end{equation}
The rationale (text sequence) of $\tau^*_a$ is then extracted to form the qualitative evidence in the \emph{consensus packet} for Phase II.

\textbf{Confidence-Weighted Voting (CWV).}
While standard MV (Eq.~\ref{eq:mv}) treats all traces as equally informative, we employ Confidence-Weighted Voting (CWV) to prioritize logically stable reasoning paths. Each trace's contribution is weighted by its prefix stability $S(\tau)$, and the final prediction is the answer with the highest aggregate support $V(a)$:
\begin{equation}
\hat{a}_{\text{CWV}}(\mathcal{T}) = \arg\max_a V(a).
\label{eq:cwv}
\end{equation}
In our experimental setup, the \emph{Online} baseline is defined as applying CWV directly to the initial stable pool $\mathcal{T}$. In contrast, our proposed PACER method applies the same CWV rule to the reasoning traces \emph{after} they have undergone the consensus-conditioned revision process described in Section~\ref{sec:methodology}.

\subsection{Token Accounting}
\label{subsec:token_accounting}

We measure inference cost as the total number of generated tokens, including:
(i) warmup full traces,
(ii) online attempts (counting tokens generated before early stopping),
and, for PACER , (iii) revision tokens and (iv) any packet-construction tokens if summarization uses the model.
Formally,
\begin{align}
\textsc{Tokens} &= \sum_{\tau\in\mathcal{T}_{\text{warm}}}|\tau|
+
\sum_{\text{attempt }i}|\tau_i^{\text{(attempt)}}|
+
\sum_{\tau\in\mathcal{T}}|\tau^{\text{(review)}}|\notag\\
&+\textsc{PacketTokens},\label{eq:token_cost}
\end{align}
where \textsc{PacketTokens}  represents the potential overhead of LLM-based summarization. In our implementation, the consensus packet is constructed via deterministic truncation of existing traces. Consequently, \textbf{$\textsc{PacketTokens} = 0$}, meaning the coordination signal is generated with zero token overhead. The only additional cost arises from the short revision step ($\tau^{\text{(review)}}$).

\section{Packet-Conditioned Revision
(PACER)}
\label{sec:methodology}

Consensus-Conditioned Revision (PACER) is a training-free coordination layer that augments token-efficient sampling with a single, low-token revision step.
PACER is designed to plug into confidence-guided inference (we instantiate it with a reproducible DeepConf-Online variant in Sec.~\ref{subsec:deepconf_variant}).
The key idea is to expose each completed trace to a compact summary of \emph{set-level evidence}—which answers are most supported and what representative rationales look like—then allow a brief self-review that can correct errors without generating many additional full traces.

% \subsection{Setup and Trace Pool Construction}
% \label{subsec:PACER_pool}

% Given a prompt $x$, we first run DeepConf-Online with $N_{\text{try}}$ total sampling attempts, including $N_{\text{init}}$ warmup traces decoded to completion.
% Warmup traces are used to estimate a stability threshold $s$ (Eq.~\eqref{eq:threshold}) and online attempts are early-stopped and discarded once their online stability drops below $s$ (Eq.~\eqref{eq:early_stop_rule}).

% Let $\mathcal{T}_{\text{warm}}$ denote the set of warmup full traces and $\mathcal{T}_{\text{surv}}$ denote the set of completed online traces that survive early stopping.
% PACER coordinates over the stable completed pool
% \begin{equation}
% \mathcal{T}
% \;=\;
% \{\tau\in\mathcal{T}_{\text{warm}} \mid S(\tau)\ge s\}
% \;\cup\;
% \mathcal{T}_{\text{surv}},
% \label{eq:PACER_pool}
% \end{equation}
% where $S(\tau)$ is the completed-trace stability (Sec.~\ref{subsec:deepconf_variant}).
% For each $\tau\in\mathcal{T}$ we extract an answer $a=\textsc{Extract}(\tau)$ and form pairs $(\tau,a)$.
% We write $B=|\mathcal{T}|$ for the realized number of completed traces; $B$ is data-dependent even when $N_{\text{try}}$ is fixed.

% \textbf{Why retain screened warmup traces?}
% Unlike variants that discard warmup traces after threshold estimation, we retain warmup traces that pass the same stability threshold.
% This increases diversity at negligible additional cost (since warmup traces were already generated) and ensures the coordination signal reflects both early variability and screened stable reasoning.

\subsection{Setup and Trace Pool Construction}
\label{subsec:PACER_pool}

Given a prompt $x$, we first run DeepConf-Online with $N_{\text{try}}$ total sampling attempts, including $N_{\text{init}}$ warmup traces decoded to completion. The warmup traces are used to estimate a stability threshold $s$ (Eq.~\eqref{eq:threshold}) and subsequent online attempts are early-stopped and discarded if their online stability $S_t(\tau)$ drops below $s$ (Eq.~\eqref{eq:early_stop_rule}). 

Let $\mathcal{T}_{\text{warm}}$ denote the set of warmup full traces and $\mathcal{T}_{\text{surv}}$ denote the set of completed online traces that survive early stopping. PACER coordinates over a stable completed pool $\mathcal{T}$, which is constructed by retaining only those traces that meet the stability requirement:
\begin{equation}
\mathcal{T}
\;=\;
\{\tau\in\mathcal{T}_{\text{warm}} \mid S(\tau)\ge s\}
\;\cup\;
\mathcal{T}_{\text{surv}},
\label{eq:PACER_pool}
\end{equation}
where $S(\tau)$ is the completed-trace stability (Sec.~\ref{subsec:deepconf_variant}). For each $\tau\in\mathcal{T}$, we extract an answer $a=\textsc{Extract}(\tau)$. We write $B=|\mathcal{T}|$ for the realized number of completed traces; $B$ is data-dependent even when $N_{\text{try}}$ is fixed.

\textbf{Dual Filtering: Online Screening and Top-$N$ Selection.} To ensure the coordination signal is both locally robust and globally representative, we implement a dual-filtering mechanism. Following the initial online screening that forms $\mathcal{T}$, we perform a second filtering step by selecting the Top-$N$ candidate answers $\mathcal{A}_{\text{top}}$. For each unique answer $a$ in the pool, we compute its confidence-weighted support $V(a) = \sum S(\tau_i)$. We then filter the distribution to retain only the $N$ answers with the highest weighted support:
\begin{equation}
\mathcal{A}_{\text{top}} = \{a^{(1)}, \dots, a^{(N)}\} \text{ s.t. } V(a^{(1)}) \ge \dots \ge V(a^{(N)}).
\label{eq:top_n_filtering}
\end{equation}
This two-stage approach ensures that the \emph{consensus packet} constructed in the next step focuses only on the most promising logical directions identified by the ensemble.

\subsection{Consensus Packet}
\label{subsec:PACER_packet}

The consensus packet is a bounded-length text summary of the pool $\mathcal{T}$ that provides low-bandwidth peer evidence.
It consists of the Top-$N$ distinct extracted answers, their support counts in the pool, and one representative compressed rationale per answer.

% \textbf{Answer support.}
% For any answer $a$, let
% $c(a) \;=\; \sum_{(\tau_i,a_i)\in\mathcal{T}} \mathbb{I}[a_i=a]$
% denote its support count.
% Let $\mathcal{A}_{\text{top}}$ be the set of the $N$ answers with largest $c(a)$.

% \textbf{Representative trace selection.}
% For each $a\in\mathcal{A}_{\text{top}}$, we select a representative trace as the most stable trace among those producing $a$:
% \begin{equation}
% \tau_a \;=\; \arg\max_{\tau:\ \textsc{Extract}(\tau)=a} S(\tau).
% \label{eq:rep_trace_PACER}
% \end{equation}
% This choice biases the packet toward coherent rationales for each candidate answer and reduces the chance that the packet amplifies unstable reasoning.
\textbf{Answer support.}
Following the definition in Sec.~\ref{subsec:pool_extraction}, we calculate the confidence-weighted support for each answer:
$V(a) \;=\; \sum_{(\tau_i,a_i)\in\mathcal{T}} S(\tau_i) \cdot \mathbb{I}[a_i=a]$.
We then select the set $\mathcal{A}_{\text{top}}$ consisting of the $N$ answers with the largest $V(a)$. 

\textbf{Representative trace selection.}
For each $a\in\mathcal{A}_{\text{top}}$, we identify the representative trace $\tau_a$ as the most stable justification (Eq.~\eqref{eq:rep_trace}). 
This choice ensures that the rationale provided in the packet for each candidate reflects the highest-confidence reasoning available in the pool.

\textbf{Rationale compression and bounded size.}
To keep packet length independent of $B$, we compress each representative rationale to a fixed budget $L_{\text{sum}}$ tokens.
In the main experiments we use deterministic truncation $\mathrm{summ}_a=\textsc{Trunc}(\tau_a;L_{\text{sum}})$, %(Sec.~\ref{subsec:packet_format}), 
and we ablate alternative compression schemes.
The consensus packet is then
\begin{equation}
\packet \;=\; \big\{(a,\ V(a),\ R_a)\big\}_{a\in\mathcal{A}_{\text{top}}}.
\label{eq:packet_PACER}
\end{equation}
Because $|\packet|=N$ and each summary is bounded, packet overhead scales as $O(N \cdot L_{\text{sum}})$ rather than $O(B)$.

\subsection{Single-Round Consensus-Conditioned Revision}
\label{subsec:PACER_revision}

Given a completed trace $(\tau_i,a_i)\in\mathcal{T}$, PACER performs a short review that conditions on the packet and outputs a revised answer $a_i'$:
$a_i' \leftarrow \textsc{Review}(x,\tau_i,a_i,\packet;\ L_{\text{rev}})$.

We enforce a strict review budget $L_{\text{rev}}$ tokens to keep the additional cost predictable.
The review prompt %(Sec.~\ref{subsec:review_prompt}) 
includes (i) the original problem, (ii) the trace's original solution and extracted answer, and (iii) the consensus packet.
The model is instructed to either keep the answer if confident or revise it if peer evidence suggests an error.

\textbf{Design rationale.}
PACER uses a \emph{single} revision step rather than multi-round interaction.
This choice keeps overhead small while enabling two error-correction mechanisms that are unavailable under independent sampling:
(i) \emph{support awareness}: a trace can reconsider when its answer has low support relative to alternatives; and
(ii) \emph{rationale contrast}: a trace can detect missing steps or contradictions by comparing its own reasoning against a representative rationale.

% \subsection{Aggregation}
% \label{subsec:PACER_agg}

% PACER outputs the final prediction by majority vote over revised answers:
% \begin{equation}
% \hat{a} =
% \arg\max_{a}\sum_{i=1}^{B}\mathbb{I}[a_i'=a].
% \label{eq:PACER_vote}
% \end{equation}

% \textbf{Optional confidence-weighted vote.}
% As an ablation, one may elicit a post-review confidence score $q_i'\in[0,1]$ and compute a weighted vote $\sum_{i:a_i'=a} q_i'$.
% We focus on unweighted voting to avoid reliance on self-reported calibration.

\subsection{Aggregation}
\label{subsec:PACER_agg}

PACER produces the final prediction by performing a confidence-weighted vote over the revised answers $\{a_i'\}_{i=1}^B$. 
Each revised answer is weighted by the prefix stability $S(\tau_i)$ of its original trace:
\vskip -0.19in
\begin{equation}
\hat{a} = \arg\max_{a}\sum_{i=1}^{B} S(\tau_i) \cdot \mathbb{I}[a_i'=a].
\label{eq:PACER_vote_final}
\end{equation}
\vskip -0.17in
By weighting the revised answers with their initial stability scores, we ensure that the final decision prioritizes revisions made by traces that were originally robust, while allowing those same traces to update their conclusions based on the global peer evidence presented in the consensus packet.

\subsection{Complexity and Token Cost}
\label{subsec:PACER_cost}

Compared to DeepConf-Online without revision, PACER adds:
(i) packet construction overhead, which is $O(N\cdot L_{\text{sum}})$ tokens if summaries are produced by the model and near-zero if summaries are obtained by truncation; and
(ii) one review completion per trace with budget $L_{\text{rev}}$, for a total of at most $B\cdot L_{\text{rev}}$ additional generated tokens.
Thus,PACER's  additional cost is bounded and typically small relative to generating hundreds of full traces in MV@256.
Algorithm~\ref{alg:PACER} summarizes the full procedure.

\begin{figure}[t]
\centering
% Keep the 20% reduction (0.75 width)
\resizebox{0.75\columnwidth}{!}{
\begin{tikzpicture}[
    node distance=0.25cm, % Tight vertical spacing to fit the extra box
    % --- STYLES ---
    process/.style={
        rectangle, 
        minimum width=3.0cm, 
        minimum height=0.45cm, 
        text centered, 
        draw=gray!30, 
        fill=white, 
        rounded corners=2pt,
        blur shadow={shadow blur steps=2, shadow xshift=0.2pt, shadow yshift=-0.2pt},
        font=\sffamily\tiny
    },
    doc/.style={
        draw=gray!40,
        fill=white,
        minimum width=0.3cm,
        minimum height=0.4cm,
        font=\tiny\sffamily,
        blur shadow={shadow blur steps=1, shadow xshift=0.2pt, shadow yshift=-0.2pt}
    },
    packet/.style={
        rectangle,
        draw=orange!60!black,
        thick,
        fill=white,
        minimum width=3.2cm,
        minimum height=0.6cm, 
        align=center,
        font=\sffamily\tiny,
        rounded corners=2pt,
        blur shadow
    },
    arrow/.style={->, >=latex, thick, color=gray!80},
    phase_label/.style={font=\tiny\bfseries\sffamily, text opacity=1.0, anchor=north east, inner sep=2pt, scale=0.8}
]

% ==========================================
% PHASE I: FILTERING
% ==========================================

\node[circle, fill=black, inner sep=1.2pt, label={[font=\tiny\bfseries, yshift=-2pt, scale=0.8]left:Prompt}] (start) {};

% Drafts
\node[doc, below=0.2cm of start, fill=gray!10] (d2) {}; 
\node[doc, left=0.08cm of d2, yshift=0.05cm, fill=gray!10] (d1) {}; 
\node[doc, right=0.08cm of d2, yshift=0.05cm, fill=gray!10] (d3) {}; 
\node[left=0.05cm of d1, font=\tiny\sffamily, color=gray, anchor=east, scale=0.7] {Traces};

% 1. DeepConf Filter
\node[process, below=0.2cm of d2, text width=3.0cm, inner sep=1pt] (filter) {
    \scalebox{0.7}{\textbf{DeepConf Filter} (Min-Conf)}
};

% 2. NEW: Top-N Selection
\node[process, below=0.2cm of filter, text width=3.0cm, inner sep=1pt] (topn) {
    \scalebox{0.7}{\textbf{Top-N Selection}} \\[-1pt]
    \scalebox{0.6}{(SumC Ranking)}
};

% Stable Pool (Moved below Top-N)
\node[doc, below=0.25cm of topn, xshift=-0.1cm, fill=blue!5, draw=blue!40] (s1) {};
\node[doc, at=(s1), xshift=0.2cm, yshift=0.05cm, fill=blue!5, draw=blue!40] (s2) {};
\node[right=0.1cm of s2, font=\tiny\bfseries, color=blue!60!black, align=left, scale=0.7] (pool_lbl) {Stable\\Pool};

% Connections Phase I
\draw[arrow] (start) -- (d2.north);
\draw[arrow] (d2) -- (filter);
\draw[arrow] (filter) -- (topn); % Connect Filter -> TopN
\draw[arrow] (topn) -- (s1);     % Connect TopN -> Pool

% ==========================================
% PHASE II: REVISION
% ==========================================

% Anchor for Phase II Box
\coordinate (p2_top) at ($(s1.south) + (0, -0.25cm)$);

% Consensus Packet
\node[packet, below=0.2cm of p2_top, xshift=0.1cm] (pkt) {
    \scalebox{0.7}{\textbf{Consensus Packet}} \\[-1pt]
    \scalebox{0.65}{ 
    \begin{tabular}{@{}l@{\hspace{3pt}}l@{}}
    \textcolor{black}{A: 45\%} & \textcolor{gray}{[Reasoning...]} \\[-1pt]
    \textcolor{black}{B: 10\%} & \textcolor{gray}{[Reasoning...]}
    \end{tabular}
    }
};

% Double Arrow Bridge
\draw[arrow, double, double distance=1.2pt] (s1.south) -- (pkt.north -| s1);

% Revision
\node[process, below=0.2cm of pkt, text width=3.0cm, inner sep=1pt] (revise) {
    \scalebox{0.7}{\textbf{Packet-Conditioned Revision}}
};

% Final Prediction
\node[process, below=0.2cm of revise, fill=blue!60!black, text=white, minimum width=2.2cm, minimum height=0.45cm] (final) {
    \scalebox{0.7}{\textbf{Final Prediction}}
};

% Connections Phase II
\draw[arrow] (pkt) -- (revise);
\draw[arrow] (revise) -- (final);

% ==========================================
% BACKGROUND BOXES
% ==========================================
\begin{pgfonlayer}{background}
    % Phase 1 Box (Expanded to include topn)
    \node[fit=(start) (d1) (d3) (pool_lbl) (s2), 
          fill=blue!5, draw=blue!20, dashed, rounded corners, inner sep=3pt] (p1) {};
    \node[phase_label, color=blue!60!black] at (p1.north east) {PHASE I};

    % Phase 2 Box
    \node[fit=(p2_top) (pkt) (final), 
          fill=orange!5, draw=orange!20, dashed, rounded corners, inner sep=3pt] (p2) {};
    \node[phase_label, color=orange!60!black] at (p2.north east) {PHASE II};
\end{pgfonlayer}

\end{tikzpicture}
}
\vspace{-3pt}
\caption{\textbf{Overview of PACER.} Noisy traces are filtered and ranked (Phase I), aggregated into a Consensus Packet, and used for revision (Phase II).}
\label{fig:ccr_pipeline}
\end{figure}
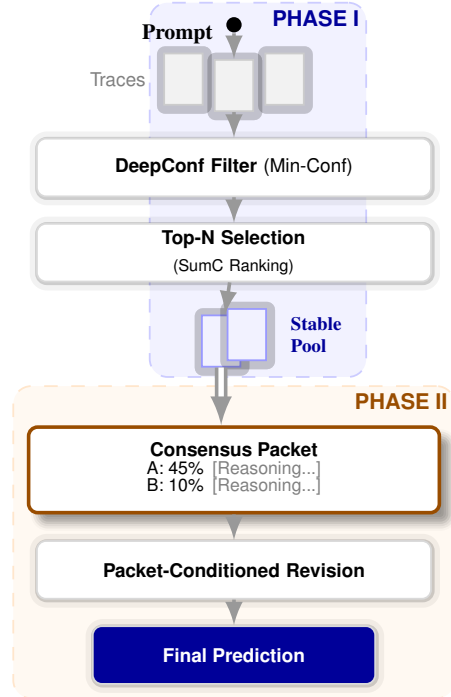

\section{Theoretical Analysis}
\label{sec:theory}

This section provides simple, checkable conditions under which a \emph{single} consensus-conditioned revision step improves accuracy, and explains why this improvement can be token-efficient relative to generating additional full traces.

\subsection{Revision as Repair vs.\ Damage}
\label{subsec:repair_damage}

Consider a trace drawn from the stable pool $\mathcal{T}$ (Eq.~\eqref{eq:PACER_pool}) with extracted pre-review answer $A$ and post-review answer $A'$ after applying $\textsc{Review}(\cdot)$ conditioned on the consensus packet $\packet$ (Eq.~\eqref{eq:packet_PACER}).
Let $y^*$ be the ground-truth answer for prompt $x$ and define the pre-review correctness probability
\[
p \;=\; \Pr(A=y^*).
\]
We summarize the effect of revision by:
\begin{itemize}
\item \textbf{Repair rate} $\alpha = \Pr(A'=y^* \mid A\neq y^*)$,
\item \textbf{Damage rate} $\beta = \Pr(A'\neq y^* \mid A=y^*)$.
\end{itemize}

\begin{assumption}[Stabilizing revision]
\label{assump:stabilizing}
Revision is stabilizing if the prevalence-adjusted mass of repairs dominates damages:
\begin{equation}
(1-p)\alpha \;\ge\; p\beta.
\label{eq:stabilizing_condition}
\end{equation}
\end{assumption}

\begin{proposition}[Per-trace accuracy gain]
\label{prop:per_trace_gain}
Under Assumption~\ref{assump:stabilizing},
\[
\Pr(A'=y^*) \;\ge\; \Pr(A=y^*),
\]
with strict inequality when \eqref{eq:stabilizing_condition} is strict.
\end{proposition}

% \begin{proof}
% By the law of total probability,
% \[
% \Pr(A'=y^*)=(1-\beta)p+\alpha(1-p),
% \]
% so
% \[
% \Pr(A'=y^*)-p=(1-p)\alpha-p\beta\ge 0.
% \]
% \end{proof}

\textbf{Why this is useful.}
Proposition~\ref{prop:per_trace_gain} is intentionally minimal, but by itself can feel tautological.
Next we tie $\alpha,\beta$ to \emph{observable packet statistics} (support counts and margins), yielding predictions that are directly testable in ablations.

\subsection{Consensus Margin and Revision Dynamics}
\label{subsec:margin_dynamics}

Let $V(a)$ be the confidence-weighted support of answer $a$ in the stable pool $\mathcal{T}$ (Eq.~2.9), and let
\(
a^{(1)} \;=\; \arg\max_a V(a)
\)
denote the leading answer in the pool.
Define the \textbf{confidence-weighted margin}
\begin{equation}
\Delta \;=\; V(a^{(1)}) - \max_{a\neq a^{(1)}} V(a),
\label{eq:margin}
\end{equation}
which measures the concentration of ensemble evidence on its leader.

We consider repair and damage rates \emph{conditioned on the ensemble evidence summarized by $\Delta$}:
$\alpha(\Delta)=\Pr(A'=y^*\mid A\neq y^*,\Delta)$,
$\beta(\Delta)=\Pr(A'\neq y^*\mid A=y^*,\Delta)$.
Intuitively, larger margins indicate clearer peer evidence and should make revision more robust.

\begin{assumption}[Margin-monotone consensus influence]
\label{assump:margin_monotone}
There exist nondecreasing $\underline{\alpha}(\Delta)$ and nonincreasing $\overline{\beta}(\Delta)$ such that for all $\Delta$,
\(
\alpha(\Delta)\ge \underline{\alpha}(\Delta),
\beta(\Delta)\le \overline{\beta}(\Delta).
\)
Moreover, for margins above a threshold $\Delta_0$, the ensemble evidence is sufficiently strong that
\begin{equation}
(1-p)\,\underline{\alpha}(\Delta) \;\ge\; p\,\overline{\beta}(\Delta)
\quad\text{for all }\Delta\ge \Delta_0.
\label{eq:margin_stabilizing}
\end{equation}
\end{assumption}

Assumption~\ref{assump:margin_monotone} is \emph{empirically checkable}: we can estimate $\alpha(\Delta)$ and $\beta(\Delta)$ from flip statistics grouped by observed margins (Section~\ref{sec:experiments}).

\begin{corollary}[Margin-conditioned per-trace improvement]
\label{cor:margin_gain}
Under Assumption~\ref{assump:margin_monotone}, for all $\Delta\ge \Delta_0$,
\[
\Pr(A'=y^*\mid \Delta)\;\ge\;\Pr(A=y^*\mid \Delta).
\]
\end{corollary}
% \begin{proof}

% Apply Proposition~\ref{prop:per_trace_gain} conditional on $\Delta$, using \eqref{eq:margin_stabilizing}.

% \end{proof}
\textbf{Interpretation.}
Corollary~\ref{cor:margin_gain} formalizes a key PACER intuition: revision is most reliably beneficial when the pool exhibits a clear leader (large $\Delta$), allowing the coordination process to effectively repair erroneous traces while minimizing damage to correct ones.

% \subsection{From Per-trace Improvement to Better Voting}
% \label{subsec:mv_amplification}

% PACER’s final prediction is CWV over revised answers (Eq.~\eqref{eq:cwv}).
% Per-trace improvements matter because CWV amplifies even small gains when the per-trace correctness exceeds $1/2$.

% Let $B=|\mathcal{T}|$ be the number of completed traces in the pool and let
% \[
% p'(\Delta) \;=\; \Pr(A'=y^*\mid \Delta).
% \]
% For analysis, we adopt a standard conditional-independence approximation: \emph{conditional on the packet (hence on $\Delta$), the randomness in review outputs across traces is independent} (e.g., due to independent sampling noise in decoding).\footnote{While the shared packet introduces dependence in the \emph{means} through $\Delta$, the conditional-independence approximation is widely used in analyses of ensembles with shared side information. We validate its qualitative predictions by measuring empirical majority accuracy as a function of $\Delta$.}

% \begin{proposition}[Majority error decays with $B$ when $p'(\Delta)>\tfrac12$]
% \label{prop:chernoff}
% Assume conditional independence given $\Delta$ and let $p'(\Delta)>\frac12$.
% Then the probability that majority vote over revised answers is incorrect satisfies
% \begin{equation}
% \Pr(\hat{a}\neq y^*\mid \Delta)
% \;\le\;
% \exp\!\big(-2B\,(p'(\Delta)-\tfrac12)^2\big).
% \label{eq:chernoff}
% \end{equation}
% \end{proposition}

\subsection{From Per-trace Improvement to Better Voting}\label{subsec:mv_amplification}PACER’s final prediction is CWV over revised answers (Eq.~\eqref{eq:cwv}).Per-trace improvements are critical because the weighted aggregation amplifies even modest accuracy gains once the per-trace correctness $p'(\Delta)$ exceeds $1/2$.Let $B=|\mathcal{T}|$ be the number of completed traces in the pool and let$$p'(\Delta) \;=\; \Pr(A'=y^*\mid \Delta).$$For analysis, we adopt a standard conditional-independence approximation: \emph{conditional on the ensemble evidence (summarized by the margin $\Delta$)}, the randomness in review outputs across traces is assumed to be independent.\footnote{While the shared textual packet introduces dependence in the \emph{means} through $\Delta$, the conditional-independence approximation is a standard tool for analyzing ensembles with shared contextual evidence. We validate its qualitative predictions by measuring empirical CWV accuracy as a function of $\Delta$.}
\begin{proposition}[CWV error decays with $B$ when $p'(\Delta)>\tfrac12$]\label{prop:chernoff}Assume conditional independence given $\Delta$ and let $p'(\Delta)>\frac12$.Then the probability that the Confidence-Weighted Vote over revised answers is incorrect satisfies\begin{equation}
\Pr(\hat{a}\neq y^*\mid \Delta)
\;\le\;
\exp\!\big(-2B\,(p'(\Delta)-\tfrac12)^2\big).
\label{eq:chernoff}
\end{equation}\end{proposition}
% \begin{proof}
% Let $Z_i=\mathbb{I}[A'_i=y^*]$ for $i=1,\dots,B$.
% Conditional on $\Delta$, the $Z_i$ are i.i.d.\ Bernoulli with mean $p'(\Delta)$.
% Majority vote fails iff $\sum_i Z_i \le B/2$.
% Apply Hoeffding’s inequality.
% \end{proof}

\textbf{Why PACER can match larger ensembles.}Proposition~\ref{prop:chernoff} shows that improving the per-trace correctness $p'(\Delta)$ through revision reduces the final ensemble error exponentially relative to the pool size $B$.Consequently, a smaller pool of coordinated traces with higher individual accuracy can effectively compete with much larger uncoordinated ensembles whose per-trace correctness remains near $1/2$.

% \textbf{Why PACER can match larger ensembles.}
% Proposition~\ref{prop:chernoff} shows that improving $p'(\Delta)$ can reduce ensemble error exponentially in $B$.
% Thus, a small pool with better per-trace correctness can compete with a much larger pool whose per-trace correctness is only slightly above $1/2$.

\subsection{Why One Revision Can Be Token-Efficient}
\label{subsec:token_efficiency_theory}

We now relate accuracy gains to token cost.
Let $L$ be the expected token length of a full trace, and let $L_{\text{rev}}\ll L$ be the enforced review budget (Sec.~\ref{subsec:PACER_revision}).
Suppose we can either:
(i) spend $\Delta T$ tokens generating additional full traces for MV, or
(ii) spend $\Delta T$ tokens performing reviews for PACER.

A key regime for test-time scaling is where MV exhibits diminishing returns as $B$ grows (the self-consistency curve flattens), while a brief review can increase per-trace correctness by a roughly constant amount on instances with informative margins (Corollary~\ref{cor:margin_gain}).

\begin{corollary}[A simple value-of-information comparison]
\label{cor:voi}
Fix a margin $\Delta\ge \Delta_0$ and suppose review increases per-trace correctness from $p(\Delta)$ to $p'(\Delta)=p(\Delta)+\delta(\Delta)$ with $\delta(\Delta)>0$ and $p'(\Delta)>\tfrac12$.
Under Proposition~\ref{prop:chernoff}, for a fixed pool size $B$, the exponent in \eqref{eq:chernoff} improves by
\vspace{-0.5em}
\begin{align*}
 2B\Big((p(\Delta)+\delta(\Delta)-\tfrac12)^2-(p(\Delta)-\tfrac12)^2\Big)\\
\;=\;
2B\,\delta(\Delta)\,(2p(\Delta)-1+\delta(\Delta)).   
\end{align*}
Thus, when $L_{\text{rev}}\ll L$, spending tokens on a single review per trace can increase the majority-vote exponent more per token than adding a small number of extra full traces, especially in the diminishing-returns regime where increasing $B$ yields only modest gains.
\end{corollary}

\textbf{Practical implication.}
Corollary~\ref{cor:voi} motivates PACER's  design choice: a bounded-cost revision step can be a better use of tokens than further increasing sample count once MV gains saturate.
In experiments, we instantiate this comparison by reporting accuracy vs.\ total generated tokens (Eq.~\eqref{eq:token_cost}) and by measuring $\alpha(\Delta),\beta(\Delta)$ and flip behavior as functions of $\Delta$.

\section{Experiments}
\label{sec:experiments}

We evaluate PACER on challenging mathematical reasoning benchmarks to test whether a single, consensus-conditioned revision step can recover the performance of large majority-vote ensembles while preserving token efficiency.

\subsection{Experimental Setup}

\textbf{Models and Datasets.}
We evaluate state-of-the-art reasoning models, including the DeepSeek-R1-Distill-Qwen series \cite{fu2025deep} and GPT OSS-120B \cite{agarwal2025gptoss} , on four competitive math benchmarks: AIME 2024, AIME 2025, HMMT 2025, and BRUMO 2025.

\textbf{Baselines.}
We compare against:
(1) \textbf{MV@256:} standard self-consistency with 256 full traces and unweighted majority vote \cite{wang2022self}.
(2) \textbf{DeepConf-Online (Top-10\%):} confidence-guided early stopping with screening percentile $\eta=10$ \cite{fu2025deep}.

\textbf{Budgets and Hyperparameters.}
We set the attempt budget to $B=256$ and use $N_{\text{init}}=64$ warmup traces. The screening percentile is $\eta=10$. This configuration yields a small number of completed survivors (typically $\sim$20--25) in addition to warmup traces that pass screening, resulting in a substantially lower total token cost than MV@256.

\subsection{Main Results}

Table~\ref{tab:main_results} summarizes accuracy. PACER consistently improves upon DeepConf-Online and matches or exceeds MV@256 despite using far fewer generated tokens.

Notably, PACER demonstrates strong generalization across difficulty levels. On the challenging \textbf{HMMT 2025} dataset with GPT-OSS-120B, PACER achieves a score of \textbf{28/30}, significantly outperforming the majority vote baseline of 25/30. Similarly, on \textbf{AIME 2025} with DeepSeek, PACER improves over standard voting by a wide margin (27/30 vs. 24/30).

% \begin{table}[ht]
% \caption{Accuracy (\%) across benchmarks. PACER uses survivors from DeepConf-Online Top-10\% screening with attempt budget $B=256$ and warmup size $N_{\text{init}}=64$.}
% \label{tab:main_results}
% \vskip 0.15in
% \begin{center}
% \begin{small}
% \begin{sc}
% \begin{tabular}{lccc}
% \toprule
% Dataset & MV@256 & DeepConf-Online (10\%) & \textbf{PACER (ours)} \\
% \midrule
% AIME 2024    & 90.00 & 90.00 & \textbf{90.00} \\
% AIME 2025    & 80.00 & 76.67 & \textbf{90.00} \\
% HMMT 2025    & 73.33 & 66.67 & \textbf{73.33} \\
% BRUMO 2025   & 93.33 & 90.00 & \textbf{93.33} \\
% \bottomrule
% \end{tabular}
% \end{sc}
% \end{small}
% \end{center}
% \vskip -0.1in
% \end{table}

\begin{table}[h]
\centering
\caption{$N=256, P=80$, Deepseek}
\label{tab:main_results}
\begin{tabular}{l|cccc}
\hline
dataset & PACER & online & mv \\ \hline
aime\_2025 & \textbf{27/30} & 26/30 & 24/30 \\
brumo\_2025 & \textbf{29/30} & 28/30 & 28/30 \\
hmmt\_2025 & \textbf{23/30} & 23/30 & 21/30 \\ \hline
\end{tabular}
\end{table}

\begin{table}[h]
\centering
\caption{$N=256, P=80$, GPT-OSS-120B}
\label{tab:main_results}
\begin{tabular}{l|cccc}
\hline
dataset & PACER & online & mv \\ \hline
aime\_2025 & \textbf{29/30} & 27/30 & 27/30 \\
brumo\_2025 & \textbf{27/30} & 26/30 & 27/30 \\
hmmt\_2025 & \textbf{28 /30} & 25/30 & 25/30 \\ \hline
\end{tabular}
\end{table}

\textbf{Discussion.}
While DeepConf effectively removes noise often matching MV, it hits a ceiling on "hard" errors where models converge on plausible distractors. PACER breaks this ceiling. By exposing survivors to divergent peer rationales, it prompts the model to switch to the correct path, yielding consistent gains (+2--3 problems) on difficult benchmarks. These results hold across architectures, suggesting Consensus-Conditioned Revision is a robust inference scaling mechanism.

\subsection{Ablation Studies}

\begin{figure}[t]
\centering
\includegraphics[width=0.7\columnwidth]{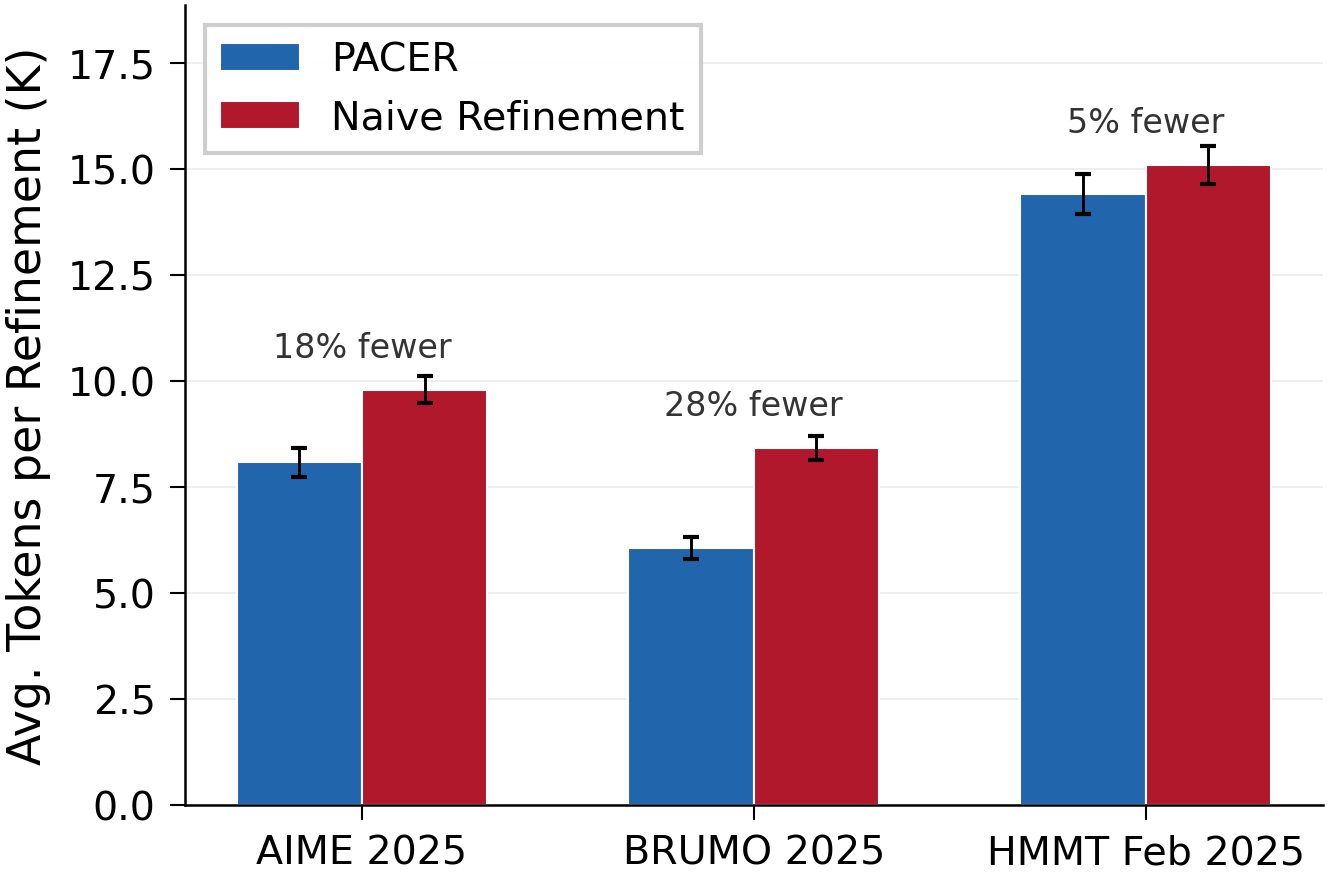}
\vspace{-5pt}
\caption{\textbf{Efficiency of directed revision.} Comparison of Phase 2 token usage between Pacer and Naive Refinement}
\label{fig:token_efficiency}
\vspace{-20pt}
\end{figure}

\textbf{Efficiency of directed revision.}
Blind self-refinement is inefficient, often requiring verbose open-ended exploration to find errors. In contrast, PACER’s consensus packet provides a \textit{directed} correction signal that allows the model to simply verify alternative logic. As shown in Figure~\ref{fig:token_efficiency}, this strategy bypasses exploration, reducing token usage by $\sim17\text{--}28\%$ compared to naive self-refinement while maintaining accuracy.

% \textbf{Packet design.}
% We vary (i) the number of candidate answers $N$ included in $\packet$, and (ii) how the representative rationale is selected (random vs. highest-stability trace). Including too many answers increases overhead and can introduce noise; we observe that small $N$ (e.g., 3--5) provides a good trade-off.

\textbf{Token efficiency.}
We measure total generated tokens (warmup + online attempts + revision prompts) and report accuracy-versus-token curves. PACER shifts the accuracy--token Pareto frontier upward relative to DeepConf-Online, showing that a short revision step can deliver accuracy gains comparable to large ensembles at a fraction of the cost.

\section*{Impact Statement}
This work studies inference-time coordination mechanisms that improve LLM reasoning accuracy under a fixed token budget.
Such methods can reduce compute requirements for legitimate applications (e.g., education and scientific assistance), but improved reasoning capability may also increase the effectiveness of misuse in domains where strong reasoning is harmful.
PACER does not introduce new training data or model weights; it is a prompting and aggregation technique.
We encourage practitioners to pair more capable inference-time methods with appropriate safety mitigations.
% and application-specific monitoring.
\newpage
\printAffiliationsAndNotice{}
\bibliographystyle{icml2026}
\bibliography{ref}
\newpage

\section{Appendix}

\subsection{Appendix A: Details of the PACER Algorithm}

In this section, we provide the detailed procedural framework for the Packet-Conditioned Revision (PACER) method. Algorithm 1 outlines the complete pipeline, including: (1) Online confidence-based screening to optimize trace quality, (2) the construction of consensus packets for top-N candidates, and (3) the final collaborative revision and aggregation process.

\begin{algorithm}[H]
\caption{Packet-Conditioned Revision (PACER)}
\label{alg:PACER}
\begin{algorithmic}[1]
\STATE {\bfseries Input:} prompt $x$; model $\theta$; total attempts $N_{\text{try}}$; warmup traces $N_{\text{init}}$; screening percentile $\eta$; Top-$N$ candidates $N$; summary budget $L_{\text{sum}}$; review budget $L_{\text{rev}}$.
\STATE {\bfseries Output:} final answer $\hat{a}$.

\vspace{0.25em}
\STATE {\bfseries 1. Online Confidence Filtering (Screening):}
\STATE Sample $\mathcal{T}_{\text{warm}}=\{\tau_j\}_{j=1}^{N_{\text{init}}}$ to completion; compute stability $\{S(\tau_j)\}$
\STATE Set threshold $s \leftarrow \textsc{Percentile}(\{S(\tau_j)\},\ 100-\eta)$
\STATE $\mathcal{T}_{\text{surv}} \leftarrow \emptyset$
\FOR{$i=1$ {\bfseries to} $N_{\text{try}}-N_{\text{init}}$}
   \STATE Decode $\tau$ token-by-token; early-stop and \textbf{discard} if $\exists\, t, S_t(\tau) < s$
   \STATE If $\tau$ completes, $\mathcal{T}_{\text{surv}} \leftarrow \mathcal{T}_{\text{surv}} \cup \{\tau\}$
\ENDFOR
\STATE $\mathcal{T} \leftarrow \{\tau \in \mathcal{T}_{\text{warm}} \mid S(\tau) \ge s\} \cup \mathcal{T}_{\text{surv}}$

\vspace{0.25em}
\STATE {\bfseries 2. Consensus Packet Construction (Top-$N$ Selection):}
\STATE For each unique $a \in \textsc{Extract}(\mathcal{T})$, compute $V(a) = \sum_{(\tau_i, a_i) \in \mathcal{T}, a_i=a} S(\tau_i)$
\STATE $\mathcal{A}_{\text{top}} \leftarrow$ Top-$N$ answers with largest $V(a)$
\STATE $\mathcal{P} \leftarrow \emptyset$
\FORALL{$a \in \mathcal{A}_{\text{top}}$}
   \STATE $\tau^*_a \leftarrow \arg\max_{\{\tau \in \mathcal{T} \mid \textsc{Extract}(\tau)=a\}} S(\tau)$ \hfill
   \STATE $R_a \leftarrow \textsc{Trunc}(\tau^*_a; L_{\text{sum}})$ \hfill 
   \STATE $\mathcal{P} \leftarrow \mathcal{P} \cup \{(a, V(a), R_a)\}$
\ENDFOR

\vspace{0.25em}
\STATE {\bfseries 3. Revision and Aggregation:}
\FORALL{$\tau_i \in \mathcal{T}$}
   \STATE $a_i \leftarrow \textsc{Extract}(\tau_i)$
   \STATE $a_i' \leftarrow \textsc{Review}(x, \tau_i, a_i, \mathcal{P}; L_{\text{rev}})$ \hfill
\ENDFOR
\STATE $\hat{a} \leftarrow \arg\max_{a} \sum_{i} S(\tau_i) \cdot \mathbb{I}[a_i'=a]$ \hfill 
\STATE {\bfseries return} $\hat{a}$
\end{algorithmic}
\end{algorithm}
% \newpage
\subsection{Appendix B: Case Study}
We provide case studies from AIME 2024 and Brumo 2025 to demonstrate PACER’s reasoning dynamics. While initial outputs fail due to local logic traps—such as misinterpreting extrema or sampling dependencies—the model successfully identifies and corrects these errors during the <refine> phase upon receiving the Consensus Packet. This highlights PACER's ability to leverage stability-weighted ensemble knowledge to shift from idiosyncratic failures to a correct global consensus through conditioned revision.

\begin{casestudybox}
\textbf{Question:} \\
There are exactly three positive real numbers $k$ such that the function
$$f(x) = \frac{(x - 18)(x - 72)(x - 98)(x - k)}{x}$$
achieves its minimum value at exactly two positive real numbers $x$. Find the sum of these values of $k$. \\

\textbf{Initial Answer (Phase I - Baseline Output):} \textcolor{red!70!black}{188} \\

\textbf{PACER Reasoning Trace:} \\
\textbf{<think>} \\
To find the minima, I analyze the critical points where $f'(x) = 0$. This is equivalent to solving:
$$\frac{1}{x-18} + \frac{1}{x-72} + \frac{1}{x-98} + \frac{1}{x-k} = \frac{1}{x}$$
If $k$ matches one of the roots $\{18, 72, 98\}$, the function has multiple roots at that point. 
Suppose $k=18$, then $x=18$ is a double root and a potential minimum. Summing $18+72+98$ gives $188$. \\
...

\textbf{<refine>} \\
\textbf{\textcolor{orange!80!black}{<Consensus Packet Received: Identifying Global Minima Conditions>}} \\
\textbf{Oops! I need to re-evaluate the "exactly two" condition.} \\
As noted in the coordination signal, if $k \in \{18, 72, 98\}$, the function's minimum value is $0$, which occurs at \textit{three} points (the roots), failing the requirement of exactly two. \\
For exactly two minima, the function must have two local minima with the \textbf{same negative value}. 
This requires the "equal-well" condition for the potential energy, which occurs at specific values of $k$ that balance the intervals. 
The valid $k$ values are determined to be 8, 32, and 200. \\
\textbf{</refine>} \\

Final sum: $8 + 32 + 200 = 240$. \\
\textbf{</think>} \\

\textbf{<answer>} \\
240 \\
\textbf{</answer>}
\end{casestudybox}
\begin{casestudybox}

\textbf{Question:} \\
Consider a $54$-deck of cards (52 + 2 jokers). Ada draws cards until she has drawn an ace, a king, and a queen. How many cards does Ada pick up on average? \\

\textbf{Initial Answer (Phase I - Baseline Output):} \\
$\frac{1354}{65} \approx 20.83$ \\

\textbf{PACER Reasoning Trace:} \\
\textbf{<think>} \\
Total relevant cards $S=12$. Expected waiting time for first success in hypergeometric: $E = \frac{N+1}{S+1}$.
Step 1: First card: $E_1 = \frac{54+1}{12+1} = \frac{55}{13}$.
Step 2: Second type: $E_2 = \frac{53+1}{8+1} = 6$.
Step 3: Third type: $E_4 = \frac{52+1}{4+1} = 10.6$. Sum: $\frac{1354}{65}$. \\

( ... more reasoning tokens omitted for brevity ) \\

\textbf{<refine>} \\
\textbf{\textcolor{orange!80!black}{<Consensus Packet Received: Validating Stage Independence in Sampling>}} \\
\textbf{Wait! My stage-based sum assumes waiting times are independent of draw position, which is false for sampling without replacement.} \\
As noted in the coordination signal, $E[W_k | T_k]$ depends on $T_k$. In a finite population, the later a stage starts, the fewer cards remain, decreasing the absolute waiting time. \\
Using inclusion-exclusion summation across $k=0$ to $54$ yields:
$$E[T] = \frac{165}{39} + \frac{220}{39} + \frac{352}{39} = \frac{737}{39} \approx 18.90.$$
The previous calculation failed to account for the diminishing pool size in later stages. \\
\textbf{</refine>} \\

Final result: $\frac{737}{39}$. \\
\textbf{</think>} \\

\textbf{<answer>} \\
$\frac{737}{39}$ \\
\textbf{</answer>}
\end{casestudybox}

\end{document}